\documentclass[11pt]{article}

\usepackage{colacl}

\title{Comparing a statistical and a rule-based tagger for German \\
{\small (Published in ``Computers, Linguistics, and Phonetics between
Language and Speech. Proc.\ of the 4th Conference on Natural Language
Processing - KONVENS-98'' Bonn, 1998. pp.125-137)}}

\author{Martin Volk and Gerold Schneider \\
University of Zurich \\
Department of Computer Science \\
Computational Linguistics Group \\
Winterthurerstr. 190, CH-8057 Zurich \\
{\tt \{volk,gschneid\}@ifi.unizh.ch}
}

\begin{document}

\maketitle

\begin{abstract}
In this paper we present the results of comparing a statistical tagger
for German based on decision trees and a rule-based Brill-Tagger for
German. We used the same training corpus (and therefore the same
tag-set) to train both taggers. We then applied the taggers to the same
test corpus and compared their respective behavior and in particular
their error rates. Both taggers perform similarly with an error rate of
around 5\%. From the detailed error analysis it can be seen that the
rule-based tagger has more problems with unknown words than the
statistical tagger. But the results are opposite for tokens that are
many-ways ambiguous. If the unknown words are fed into the taggers with
the help of an external lexicon (such as the Gertwol system) the error
rate of the rule-based tagger drops to 4.7\%, and the respective rate of
the statistical taggers drops to around 3.7\%. Combining the taggers by
using the output of one tagger to help the other did not lead to any
further improvement. 

\vspace{0.5cm}

In diesem Beitrag beschreiben wir die Resultate aus unserem Vergleich
eines statistischen Taggers, der auf Entscheidungsb\"aumen basiert, und
eines regel-basierten Brill-Taggers f\"ur das Deutsche. Beim Vergleich
benutzten wir dasselbe Trainingskorpus (und damit dasselbe Tagset), um
beide Tagger zu trainieren. Danach wurden beide Tagger auf dasselbe
Testkorpus angewendet und ihr jeweiliges Verhalten und ihre Fehlerraten
verglichen. Beide Tagger liegen ungef\"ahr bei 5\% Fehlerrate. Bei der
detaillierten Fehleranalyse sieht man, dass der regel-basierte Tagger
gr\"ossere Probleme bei unbekannten Wortformen hat als der statistische
Tagger. Bei vielfach ambigen Wortformen ist das Ergebnis jedoch
umgekehrt. Wenn man die unbekannten Wortformen mit Hilfe eines externen
Lexikons (z.B.\ mit dem Gertwol-System) reduziert, sinkt die Fehlerrate
des regel-basierten Taggers auf 4,7\% und die entsprechende Rate des
statistischen Taggers auf 3,7\%. Eine Kombination der Tagger, der Output
des einen als Hilfestellung f\"ur den anderen, brachte keine weitere
Verbesserung.
\end{abstract}

\section{Introduction}
In recent years a number of part-of-speech taggers have been developed
for German. \cite{Lezi96} list 6 taggers (all of which work with
statistical methods) and provide comparison figures. They report that
for a ``small'' tagset the accuracy of these 6 taggers varies from
92.8\% to 97\%. But these figures do not tell us much about the
comparative behavior of the taggers since the figures are based on
different tagsets, different training corpora, and different test
corpora. A more rigorous approach to comparison is necessary to obtain
valid results. Such an approach has been presented by \cite{Teuf96}.
They have developed an elaborate methodology for comparing taggers
including tagger evaluation, tagset evaluation and text type evaluation. 

\begin{description}
\item[Tagger evaluation] 
     Tests allowing to assess the impact of different tagging methods,
by comparing the performance of different taggers on the same training
and test data, using the same tagset.
\item[Tagset evaluation]
     Tests allowing to assess the impact of tagset modifications on the
results, by using different versions of a given tagset on the same
texts.  
\item[Text type evaluation] 
     Tests allowing to assess the impact of linguistic differences
between training texts and application texts, by using texts from
different text types in training and testing, tagsets and taggers being
unchanged otherwise. 
\end{description}

In this paper we will focus on ``Tagger evaluation'' for the most part,
and only in section \ref{TextTypeEval} will we briefly sidestep to
``Text type evaluation''.  

\cite{Teuf96} used their methodology only on two statistical taggers for
German, the Xerox HMM tagger \cite{Cutt92} and the TreeTagger
\cite{Schm95,Schm96}. On contrast, we will compare one of these
statistical taggers, the TreeTagger, to a rule-based tagger for German,
the Brill-Tagger \cite{Bril92,Bril94a}. Such a comparison is worthwhile
since \cite{Samu97} have shown for English that their rule-based tagger,
a constraint grammar tagger, outperforms any known statistical tagger.

\section{Our Tagger Evaluation}
For our evaluation we used a manually tagged corpus of around 70'000
tokens which we obtained from the University of
Stuttgart.\footnote{Thanks to Uli Heid for making this corpus available
to us.} The texts in that corpus are taken from the Frankfurter
Rundschau, a daily newspaper. We split the corpus into a $7/8$ training
corpus (60'710 tokens) and a $1/8$ test corpus (8'887 tokens) using a
tool supplied by Eric Brill that divides a corpus sentence by sentence.
The test corpus then contains sentences from many different sections of
the corpus. The average rate of ambiguity in the test corpus is 1.50.
That means that on average for any token in the test corpus there is a
choice of 1.5 tags in the lexicon, if the token is in the lexion.  1342
tokens from the test corpus are not present in the training corpus and
are therefore not in the lexicon (these are called ``lexicon gaps'' by
\cite{Teuf96}).

The corpus is tagged with the STTS, the Stuttgart-T\"ubingen TagSet
\cite{Schi95,Thie96}. This tagset consists of 54 tags, including 3 tags
for punctuation marks. We modified the tagset in one little aspect. The
STTS contains one tag for both digit-sequence numbers (e.g.\ {\it 2, 11,
100}) and letter-sequence numbers ({\it two, eleven, hundred}). The tag
is called CARD since it stands for all cardinal numbers. We added a new
tag, CARDNUM, for digit-sequence numbers and restricted the use of CARD
to letter-sequence numbers. The assumption was that this move makes it
easier for the taggers to recognize unknown numbers, most of which will
be digit-sequence numbers.  

\subsection{Training the TreeTagger}

In a first phase we trained the TreeTagger with its standard parameter
settings as given by the author of the tagger.\footnote{These parameters
are explained in the README file that comes with the tagger.} That is,
it was trained with 

\begin{enumerate}
\item Context length set to 2 (number of preceding words forming the
tagging context). Context length 2 corresponds to a trigram context. 
\item Minimal decision tree gain set to 0.7. If the information gain at
a leaf node of the decision tree is below this threshold, the node is
deleted.
\item Equivalence class weight set to 0.15. This weight of the
equivalence class is based on probability estimates. 
\item Affix tree gain set to 1.2. If the information gain at a leaf of
an affix tree is below this threshold, it is deleted. 
\end{enumerate}

The training took less than 2 minutes and created an output file of 630
kByte. Using the tagger with this output file to tag the test corpus
resulted in an error rate of 4.73\%. Table \ref{SchmidResults} gives an
overview of the errors.

\begin{table*}\begin{center}
\begin{tabular}{|r|rr||rr|rr|rr|}
\hline
ambiguity & tokens & in \% & correct & in \% & LE & in \% & DE & in \%
\\
\hline \hline
  0 &    1342 &  15.10 &    1128 &  84.05 &     214 &  15.95 &       0
&   0.00 \\
  1 &    5401 &  60.77 &    5330 &  98.69 &      71 &   1.31 &       0
&   0.00 \\
  2 &     993 &  11.17 &     929 &  93.55 &       3 &   0.30 &      61
&   6.14 \\
  3 &     795 &   8.95 &     757 &  95.22 &       0 &   0.00 &      38
&   4.78 \\
  4 &     260 &   2.93 &     240 &  92.31 &       0 &   0.00 &      20
&   7.69 \\
  5 &      96 &   1.08 &      83 &  86.46 &       0 &   0.00 &      13
&  13.54 \\
\hline
total &    8887 & 100.00 &    8467 &  95.27 &     288 &   3.24 &     132
&   1.49 \\
\hline
\end{tabular}\end{center}
\caption{Error statistics of the TreeTagger}
\label{SchmidResults}
\end{table*}

Column 1 lists the ambiguity rates, i.e.\ the number of tags available
to a token according to the lexicon. Note that the lexicon was built
solely on the basis of the training corpus. From columns 1 and 2 we
learn that 1342 tokens from the test corpus were not in lexicon, 5401
tokens in the test corpus have exactly one tag in the lexicon, 993
tokens have two tags in the lexicon and so on. Column 3, labelled
`correct', displays the number of tokens correctly tagged by the
TreeTagger. It is obvious that the correct assignment of tags is most
difficult for tokens that are not in the lexicon (84.05\%) and for
tokens that are many ways ambiguous (86.46\% for tokens that are 5-ways
ambiguous). 

The errors made by the tagger can be split into lexical errors (LE;
column 4) and disambiguation errors (DE; column 5). Lexical errors occur
when the correct tag is not available in the lexicon. All errors for
tokens not in the lexicon are lexical errors (214). In addition there
are a total of 74 lexical errors in the ambiguity rates 1 and 2 where
the correct tag is not in the lexicon. On the contrary, disambiguation
errors occur when the correct tag is available but the tagger picks the
wrong one. Such errors can only occur if the tagger has a choice among
at least two tags. Thus we get a rate of 3.24\% lexical errors and
1.49\% disambiguation errors adding up to the total error rate of
4.73\%.

It should be noted that this error rate is higher than the error rate
given for the TreeTagger in \cite{Teuf96}. There, the TreeTagger had
been trained over 62'860 tokens and tested over 13'416 tokens of a
corpus very similar to ours (50'000 words from the Frankfurter Rundschau
plus 25'000 words from the Stuttgarter Zeitung). \cite{Teuf96} report on
an error rate of only 3.0\% for the TreeTagger. It could be that they
were using different training parameters, these are not listed in the
paper. But more likely they were using a more complete lexicon. They
report on only 240 lexicon gaps among the 13'416 test tokens.

\subsection{Training the Brill-Tagger}

In parallel with the TreeTagger we trained the Brill-Tagger with our
training corpus using the following parameter settings. Since we had
some experience with training the Brill-Tagger we set the parameters
slightly different from the Brill's suggestions.\footnote{The
suggestions for the tagging parameters of the Brill-Tagger are given in
the README file that is distributed with the tagger.}

\begin{enumerate}
\item The threshold for the best found lexical rule was set to 2. The
learner terminates when the score of the best found rule drops below
this threshold. (Brill suggests 4 for a training corpus of 50K-100K
words.)
\item The threshold for the best found contextual rule was set to 1. The
learner terminates when the score of the best found rule drops below
this threshold. (Brill suggests 3 for a training corpus of 50K-100K
words.)
\item The bigram restriction value was set to 500. This tells the rule
learner to only use bigram contexts where one of the two words is among
the 500 most frequent words. A higher number will increase the accuracy
at the cost of further increasing the training time. (Brill suggests
300.)
\end{enumerate}

Training this tagger takes much longer than training the TreeTagger. Our
training step took around 30 hours (!!) on a Sun Ultra-Sparc
workstation. It resulted in:

\begin{enumerate}
\item a fullform lexicon with 14'147 entries (212 kByte)
\item a lexical-rules file with 378 rules (9 kByte)
\item a context-rules file with 329 rules (8 kByte)
\item a bigram list with 42'279 entries (609 kByte)
\end{enumerate}

Using the tagger with this training output to tag the test corpus
resulted in an error rate of 5.25\%. Table \ref{BrillResults} gives an
overview of the errors.

\begin{table*}\begin{center}
\begin{tabular}{|r|rr||rr|rr|rr|}
\hline
ambiguity & tokens & in \% & correct & in \% & LE & in \% & DE & in \%
\\
\hline \hline
  0 &    1342 &  15.10 &    1094 &  81.52 &     248 &  18.48 &       0
&   0.00 \\

  1 &    5401 &  60.77 &    5330 &  98.69 &      71 &   1.31 &       0
&   0.00 \\
  2 &     993 &  11.17 &     906 &  91.24 &       3 &   0.30 &      84
&   8.46 \\
  3 &     795 &   8.95 &     758 &  95.35 &       0 &   0.00 &      37
&   4.65 \\
  4 &     260 &   2.93 &     245 &  94.23 &       0 &   0.00 &      15
&   5.77 \\
  5 &      96 &   1.08 &      87 &  90.62 &       0 &   0.00 &       9
&   9.38 \\
\hline

total &    8887 & 100.00 &    8420 &  94.75 &     322 &   3.62 &     145
&   1.63 \\

\hline
\end{tabular}\end{center}
\caption{Error statistics of the Brill-Tagger}
\label{BrillResults}
\end{table*}

It is striking that the overall result is very similar to the
TreeTagger. A closer look reveals interesting differences. The
TreeTagger is clearly better than the Brill-Tagger in dealing with
unknown words (i.e.\ tokens not in the lexicon). There, the TreeTagger
reaches 84.05\% correct assignments which is 2.5\% better than the
Brill-Tagger. On the opposite side of the ambiguity spectrum the
Brill-Tagger is superior to the TreeTagger in disambiguating between
highly ambiguous tokens. For 4-way ambiguous tokens it reaches 94.23\%
correct assignments (a plus of 1.9\% over the TreeTagger) and even for
5-way ambiguous tokens it still reaches 90.62\% correct tags which is
4.1\% better than the TreeTagger. 

\subsection{Error comparison}
We then compared the types of errors made by both taggers. An error type
is defined by the tuple {\tt (correct tag, tagger tag)}, where {\tt
correct tag} is the manually assigned tag and {\tt tagger tag} is the
automatically assigned tag. Both taggers produce about the same number
of error types (132 for the TreeTagger and 131 for the Brill-Tagger).
Table \ref{ErrorTypes} lists the most frequent error types for both
taggers. The biggest problem for both taggers is the distinction between
proper nouns (NE) and common nouns (NN). This corresponds with the
findings in \cite{Teuf96}. The distribution of proper and common nouns
is very similar in German and is therefore difficult to distinguish by
the taggers.

\begin{verbatim}
er wollte auch Weber/NN?/NE? einstellen
\end{verbatim}

\begin{table*}\begin{center}
\begin{tabular}{|r|l|l||r|l|l|}
\hline
\multicolumn{3}{|c||}{TreeTagger errors} &
\multicolumn{3}{|c|}{Brill-Tagger errors} \\
\hline
number & correct tag & tagger tag & number & correct tag & tagger tag \\
\hline
48 &     NE & NN                & 54 &   NE & NN \\
21 &     VVINF & VVFIN          & 31 &   NN & NE \\
20 &     NN & NE                & 19 &   VVFIN & VVINF \\
17 &     VVFIN & VVINF          & 19 &   VVFIN & ADJA  \\
10 &     VVPP & VVFIN           & 17 &   VVINF & VVFIN \\
10 &     VVFIN & VVPP           & 15 &   VVPP & VVFIN \\
8 &      CARDNUM & VMPP         & 11 &   VVPP & ADJD \\
7 &      ADJD & VVFIN           & 11 &   ADJD & VVFIN \\
7 &      ADJD & ADV             & 8 &    VVINF & ADJA \\
\hline
\end{tabular}\end{center}
\caption{Most frequent error types}
\label{ErrorTypes}
\end{table*}

The second biggest problem results from the distinction between
different forms of full verbs: finite verbs (VVFIN), infinite verbs
(VVINF), and past participle verbs (VVPP). This problem is caused by the
limited `window size' of both taggers. The TreeTagger uses trigrams for
its estimations, and the Brill-Tagger can base its decisions on up to
three tokens to the right and to the left. This is rather limited if we
consider the possible distance between the finite verb (in second
position) and the rest of the verb group (in sentence final position) in
German main clauses. In addition, the taggers cannot distinguish between
main and subordinate clause structure.

\begin{verbatim}
... weil wir die Probleme schon kennen/VVFIN.
Wir sollten die Probleme schon kennen/VVINF.
\end{verbatim}

A third frequent error type arises between verb forms and adjectives
(ADJA: adjective used as an attribute, inflected form; ADJD: adjective
in predicative use, typically uninflected form). It might be surprising
that the Brill-Tagger has so much difficulty to tell apart a finite verb
and an inflected adjective (19 errors). But this can be explained by
looking at the lexical rules learned by this tagger. These rules are
used by the Brill-Tagger to guess a tag for unknown words
\cite{Bril94a}. And the first lexical rule learned from our training
corpus says that a word form ending in the letter {\tt e} should be
treated as an adjective (ADJA). Of course this assignment can be
overridden by other lexical rules or contextual rules, but these
obviously miss some 19 cases.

On the other hand it is surprising that the TreeTagger gets mixed up 8
times by past participle modal verbs (VMPP) which should be
digit-sequence cardinal numbers (CARDNUM). There are 10 additional cases
where a digit-sequence cardinal number was interpreted as some other tag
by the TreeTagger. But there are only 3 similar errors for the
Brill-Tagger since its lexical rules are well suited to recognize
unknown digit-sequence numbers.

\section{Using an external lexicon}
Let us sum up the results of the above comparison and see if we can
improve tagging accuracy by using an external lexicon. The above
comparison showed that:

\begin{enumerate}
\item The Brill-Tagger is better in recognizing special symbol items
such as digit-sequence cardinal numbers, and it is better in
disambiguating tokens which are many-ways ambiguous in the lexicon.
\item The TreeTagger is better in dealing with unknown word forms.
\end{enumerate}

At first sight it seems easiest to improve the Brill-Tagger by reducing
its unknown word problem. We employed the Gertwol system \cite{Ling94} a
wide-coverage morphological analyzer to fill up the tagger lexicon
before tagging starts. That means we extracted all unknown word
forms\footnote{Unknown word forms in the test corpus are all tokens not
seen in the training corpus.} from the test corpus and had Gertwol
analyse them. From the 1342 unknown tokens we get 1309 types which we
feed to  Gertwol. Gertwol is able to analyse 1205 of these types.
Gertwol's output is mapped to the respective tags, and every word form
with all possible tags is added temporarily to the tagger lexicon. In
this way the tagger starts tagging the test corpus with an almost
complete lexicon. The remaining lexicon gaps are the few words Gertwol
cannot analyse. In our test corpus 109 tokens remain unanalysed. 

Our experiments showed a slight improvement in accuracy (about 0.5\%),
but by far not as much as we had expected. The alternative of filling up
the tagger lexicon by training over the whole corpus resulted in an
improvement of around 3.5\%, an excellent tagger accuracy of more than
98\%. Note that we only used the lexicon filled in this way but the
rules as learned from the training corpus alone. But, of course, it is
an unrealistic scenario to know in advance (i.e.\ during tagger
training) the text to be tagged.

The difference between using a large external `lexicon' such as Gertwol
and using the internal vocabulary is due to two facts. First, Gertwol
increases the average ambiguity of tokens since it gives every possible
tag for a word form. The internal vocabulary will only provide the tag
occuring in the corpus. Second, in case of multiple tags for a word form
the Brill-Tagger needs to know the most likely tag. This is very
important for the Brill-Tagger algorithm. But Gertwol gives all possible
tags in an arbitrary order. One solution is to sort Gertwol's output
according to overall tag probabilities. These can be computed from the
frequencies of every tag in the training corpus irrespective of the word
form. Using these rough probabilities improved the results in our
experiments by about 0.2\%. This means that the best result for
combining Gertwol with the Brill-Tagger is at 95.45\% accuracy.

In almost the same way we can use the external lexicon with the
TreeTagger. We add all types as analysed by Gertwol to the TreeTagger's
lexicon. Then, unlike the Brill-Tagger, the TreeTagger is retrained with
the same parameters and input files as above except for the extended
lexicon. The Brill-Tagger loads its lexicon for every tagging process,
and the lexicon can therefore be extended without retraining the tagger.
The TreeTagger, on the other hand, integrates the lexicon during
training into its 'output file'. It must therefore be retrained after
each lexicon extension. 

Extending the lexicon improves the TreeTagger's accuracy by around 1\%
to 96.29\%. Table \ref{ExtendedLexResults} gives the results for the
TreeTagger with the extended lexicon.

\begin{table*}\begin{center}
\begin{tabular}{|r|rr||rr|rr|rr|}
\hline
ambiguity & tokens & in \% & correct & in \% & LE & in \% & DE & in \%
\\
\hline \hline
  0 &     109 &   1.23 &      72 &  66.06 &      37 &  33.94 &       0
&   0.00 \\
  1 &    6307 &  70.97 &    6209 &  98.45 &      98 &   1.55 &       0
&   0.00 \\
  2 &    1224 &  13.77 &    1119 &  91.42 &      10 &   0.82 &      95
&   7.76 \\
  3 &     852 &   9.59 &     805 &  94.48 &       2 &   0.23 &      45
&   5.28 \\
  4 &     296 &   3.33 &     266 &  89.86 &       0 &   0.00 &      30
&  10.14 \\
  5 &      99 &   1.11 &      86 &  86.87 &       0 &   0.00 &      13
&  13.13 \\
\hline
total &    8887 & 100.00 &    8557 &  96.29 &     147 &   1.65 &     183
&   2.06 \\
\hline
\end{tabular}\end{center}
\caption{Error statistics of the TreeTagger with an extended lexicon}
\label{ExtendedLexResults}
\end{table*}

The recognition of the remaining unknown words is very low (66.06\%),
but this does not influence the result much since only 1.23\% of all
tokens are left unknown. Also the rate of disambiguation errors
increases from 1.49\% to 2.06\%. But at the same time the rate of
lexical error drops from 3.24\% to 1.65\%, which accounts for the
noticeable increase in overall accuracy.

\section{The best of both worlds?}
In the previous sections we observed that the statistical tagger and the
rule-based tagger show complementary strengths. Therefore we
experimented with combining the statistical and the rule-based tagger in
order to find out whether a combination would yield a result superior to
any single tagger.

First, we tried to employ the TreeTagger and the Brill-Tagger in this
order. Tagging the test corpus now works in two steps. In step one, we
tag the test corpus with the TreeTagger. We then add all unknown word
forms to the Brill-Tagger's lexicon with the tags assigned by the
TreeTagger. In step two, we tag the test corpus with the Brill-Tagger.
In this way we can increase the Brill-Tagger's accuracy to 95.13\%. But
the desired effect of combining the strengths of both taggers in order
to build one tagger that is better than either of the taggers alone was
not achieved. The reason is that the wrong tags of the TreeTagger were
carried over to the Brill-Tagger (together with the correct tags) and
all of the new lexical entries were on the ambiguity level one or two,
so that the Brill-Tagger could not show its strength in disambiguation.

In a second round we reduced the export of wrong tags from the
TreeTagger to the Brill-Tagger. We made sure that on export all
digit-sequence ordinal and cardinal numbers were assigned the correct
tags. We used a regular expression to check each word form. In addition,
we checked for all other unknown word forms if the tag assigned by the
TreeTagger was permitted by Gertwol (i.e.\ if the TreeTagger tag was one
of Gertwol's tags). If so, the TreeTagger tag was exported. If the
TreeTagger tag was not allowed by Gertwol, we checked how many tags
Gertwol proposes. If Gertwol proposes exactly one tag this tag was
exported, in all other cases no tag was exported. In this way we
exported 1171 types to the Brill-Tagger's lexicon and we obtained a
tagging accuracy of 95.90\%. The algorithm for selecting TreeTagger tags
was further modified in one little respect. If Gertwol did not analyse a
word form and the TreeTagger identified it as a proper noun (NE), then
the tag was exported. We then export 1212 types and we obtain a tagging
accuracy of 96.03\%, which is still slightly worse than the TreeTagger
with the external lexicon.  
 
Second, we tried to employ the taggers in the reverse order:
Brill-Tagger first, and then the TreeTagger, using the Brill-Tagger's
output. In this test we extended the TreeTaggers lexicon with the tags
assigned by the Brill-Tagger and we extended the training corpus with
the test corpus tagged by the Brill-Tagger. We retrained the TreeTagger
with the extended lexicon and the extended corpus. We then used the
TreeTagger to tag the test corpus, which resulted in 95.05\% accuracy.
This means that the combination of the taggers results in a worse result
than the TreeTagger by itself (95.27\%). 

>From these tests we conclude that it is not possible to improve the
tagging result by simply sequentialising the taggers. In order to
exploit their respective strengths a more elaborate intertwining of
their tagging strategies will be necessary.

\section{Text type evaluation}\label{TextTypeEval}

So far, all our tests were performed over the same test corpus. We
checked whether the general tendency will also carry over to other test
corpora. Besides the corpus used for the above evaluation we have a
second manually tagged corpus consisting of texts about the
administration at the University of Zurich (the university's annual
report; guidelines for student registration etc.). This corpus currently
consists of 38'007 tokens. We have applied the taggers, trained as above
on $7/8$ of the 'Frankfurter Rundschau' corpus, to this corpus and
compared the results. In this way we have a much larger test corpus but
we have a higher rate of unknown words (10'646 tokens, 28.01\%, are
unknown). The TreeTagger resulted in an accuracy rate of 92.37\%,
whereas the Brill-Tagger showed an accuracy rate of 91.65\%. These
results correspond very well with the above findings. The figures are
close to each other with a small advantage for the TreeTagger. It should
be noted that the much lower accuracy rates compared to the test corpus
are in part due to inconsistencies in tagging decisions. E.g.\ the word
`Management' was tagged as a regular noun (NN) in the training corpus
but as foreign material (FM) in the University of Zurich test corpus.

\section{Conclusions}
We have compared a statistical and a rule-based tagger for German. It
turned out that both taggers perform on the same general level, but the
statistical tagger has an advantage of about 0.5\% to 1\%. A detailed
analysis shows that the statistical tagger is better in dealing with
unknown words than the rule-based tagger. It is also more robust in
using an external lexicon, which resulted in the top tagging accuracy of
96.29\%. The rule-based tagger is superior to the statistical tagger in
disambiguating tokens that are many-ways ambiguous. But such tokens do
not occur frequently enough to fully get equal with the statistical
tagger. A sequential combination of both taggers in either order did not
show any improvements in tagging accuracy.

The statistical tagger is easier to handle in that its training time is
3 magnitudes shorter than the rule-based tagger (minutes vs.\ days). But
it has to be retrained after lexicon extension, which is not necessary
with the rule-based tagger. The rule-based tagger has the additional
advantage that rules (i.e.\ lexical and contextual rules) can be
manually modified. As a side result our experiments show that a
rule-based tagger that learns its rules like the Brill-Tagger does not
match the results of the constraint grammar tagger (a manually built
rule-based tagger) described in \cite{Samu97}. That tagger is described
as performing with an error rate of less than 2\%. Constraint grammar
rules are much more powerful than the rules used in the Brill-Tagger.

\end{document}